\documentclass{article}

\usepackage{arxiv}

\usepackage[utf8]{inputenc} 
\usepackage[T1]{fontenc}    
\usepackage{hyperref}       
\usepackage{url}            
\usepackage{booktabs}       
\usepackage{amsfonts}       
\usepackage{nicefrac}       
\usepackage{microtype}      
\usepackage{lipsum}
\usepackage{graphicx}
\usepackage[normalem]{ulem}
\usepackage{pythonhighlight}
\graphicspath{ {./images/} }

\newcommand\SALINA{\texttt{SaLinA}}
\title{SaLinA: Sequential Learning of Agents}

\author{
 Ludovic Denoyer, Alfredo de la Fuente, Song Duong, Jean-Baptiste Gaya, \\
 \textbf{Pierre-Alexandre Kamienny, Daniel H. Thompson} \\
  Facebook \\
  \texttt{denoyer@fb.com} \\
  Github: \texttt{https://github.com/facebookresearch/salina} \\
}

\begin{document}
\maketitle
\begin{abstract}
\SALINA{} is a simple library that makes implementing complex sequential learning models easy, including reinforcement learning algorithms. It is built as an extension of PyTorch \cite{pytorch}: algorithms coded with \SALINA{} can be understood in few minutes by PyTorch users and modified easily. Moreover, \SALINA{} naturally works with multiple CPUs and GPUs at train and test time, thus being a good fit for the large-scale training use cases. In comparison to existing RL libraries, \SALINA{} has a very low adoption cost and capture a large variety of settings (model-based RL, batch RL, hierarchical RL, multi-agent RL, etc.). But \SALINA{} does not only target RL practitioners, it aims at providing sequential learning capabilities to any deep learning programmer.

\end{abstract}

\tableofcontents
\newpage

\section{Context and Motivations}
Many Deep Learning (DL) libraries (e.g PyTorch, JAX, Tensorflow) have been conceived to facilitate the implementation of complex differentiable functions (aka neural networks). The typical shape of such functions is $f(x) \rightarrow y$ where $x$ is a (set) of input tensors, and $y$ is a (set) of output tensors produced by executing multiple computations over $x$. Implementing a new architecture (i.e a new $f$ function) is made by assembling multiple blocks (or modules) through composition operators, allowing one to quickly prototype any new idea. But this approach does not well handle the implementation of sequential decision methods. Indeed, classical platforms do not provide efficient ways to naturally write $f$ as a sequential (possibly stochastic) process where the information is acquired, processed, and transformed step by step.

This is particularly critical when implementing Reinforcement Learning (RL) algorithms: if specifying how an agent processes observations at time $t$ can be handled with classical deep learning framework, modeling the interaction of the agent with the environment over multiple steps requires writing extra-code which does not naturally integrate with the DL platforms principles. As a solution, multiple specific RL frameworks have been developed in the past few years \cite{rllib,acme,baselines,mtrl}, but suffer from two main drawbacks: 
\begin{itemize}
\item They usually define (many) new (complex) abstractions and are, in philosophy, very far from the DL framework they rely on. As a consequence, they have a high adoption cost, low flexibility, and are difficult to use, particularly for people not familiar with reinforcement learning. 
\item Moreover, these platforms are usually specific to RL and to specific cases of RL (e.g model-based RL, batch-RL, multi-agent RL, etc...). There is, as far as we know, no RL library that can naturally handle all RL settings, and no library that naturally capture the RL setting, but that also can be used to implement sequential learning models in general like cascade models or attention-models, etc..
\end{itemize}

\SALINA{} is an attempt to solve these two problems: it pursues the objective of making the implementation of sequential decision processes (including RL methods) as simple and natural as implementing neural network architectures. For that, it proposes to implement any sequential decision problem through the composition of simple 'agents' that process information sequentially. The targeted audience is not only RL researchers but also computer vision researchers aiming at giving a sequential decision dimension to their methods, NLP researchers looking for a natural way to model dialog, etc... \SALINA{} is built as an extension of PyTorch\ (but the principles could be extended to other DL libraries like JAX) and has a core code of few hundred lines that is easy to understand and maintain.

We now describe the founding principles of the library and explain how \SALINA{} can be used to implement complex algorithms. Examples of code are shown in Section \ref{section:examples}, but the library also comes with tutorials, videos, and example algorithms, particularly in the RL domain. It is available at \url{https://github.com/facebookresearch/salina}.

\section{Principles}

DL frameworks allow one to define a function $f(x) \rightarrow y$ by composing modules or operators. In PyTorch it is made by using the \texttt{torch.nn.Module} class where the \texttt{forward(self,x)} function has to be implemented by the programmer to compute a particular transformation. Multiple modules can then be combined to generate more complex functions, by chaining them for instance. Moreover, PyTorch comes with many pre-implemented modules to facilitate the development of new models. \SALINA{} keeps the same philosophy. It is built as an extension of PyTorch , adding a few principles that together are a game-changer for implementing sequential learning systems.

\paragraph{Principle 1: All \sout{modules} agents exchange information through a Workspace.}  \SALINA{} defines a \texttt{salina.Workspace} object allowing to organize tensors. Basically, a workspace\footnote{A workspace can be seen as an implementation of a black board system (see \url{https://en.wikipedia.org/wiki/Blackboard_system})} is like a dictionary of tensors where each tensor has a time dimension and a batch dimension, but provides additional high-level functions to manipulate these tensors. A workspace is used to store complex temporal traces, either generated by the model, loaded from a dataset, etc. (see Figure \ref{code:workspace}). 

Given a workspace, the first principle of \SALINA{} is to define agents that will exchange information through a workspace, by reading and writing variables in/from it.

Concretely, the \texttt{torch.nn.Module} is extended through the  \texttt{salina.Agent} class\footnote{\texttt{salina.Agent} inherits from \texttt{torch.nn.Module} and thus has all the capabilities of classical PyTorch modules, like moving to different devices, generating a list of parameters, etc.}. To define an \texttt{Agent}, the user overrides the \texttt{Agent.forward} function (see Figures \ref{code:linearagent}, \ref{code:lossagent} for definition of agents, and Figure \ref{code:executeagent} for an example of execution of an agent). Note that the workspace is implicit when defining the \texttt{forward} function, the agent accessing the workspace through the \texttt{self.set} and \texttt{self.get} methods and does not appear as an argument of the \texttt{forward} function. This seems to be a small change in the PyTorch API, but it reveals to be one of the key principles that greatly simplifies the writing of complex sequential agents: each agent has access to all the information produced by other agents at any time step. As shown later, this principle allows one to develop learning algorithms that are independent of the underlying architecture of the agents (for instance, all RL algorithms provided in \SALINA{} works with classical NNs, recurrent NNs, transformers, etc): if two agents are reading/writing the same variables, they can be easily exchanged at compute time. 

\begin{figure}[t!]
\begin{center}
    \includegraphics[width=0.8\textwidth]{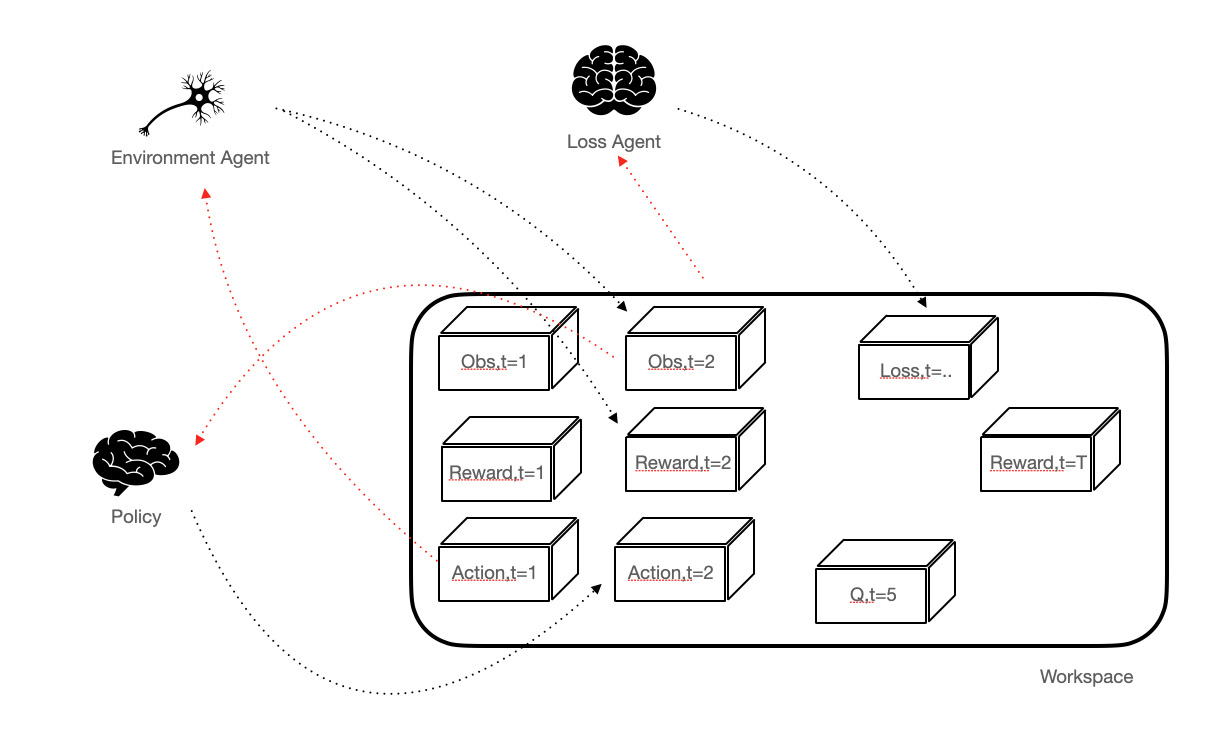}
\end{center}
\caption{\SALINA{} principles: Multiple agents read (black dotted lines) and write (red dotted lines) information into a workspace. This workspace corresponds to a common knowledge base which is iteratively updated by a diverse group of agents. Very similar to a blackboard in multi-agent systems, this architecture handles complex, ill-defined problems, where the solution is the sum of its parts.}
\end{figure}

\paragraph{Principle 2: Everything is an Agent. } All the components used in \SALINA{} are  \texttt{Agent}s. As an example, in Reinforcement Learning, an environment is an \texttt{Agent} that reads actions (at time $t-1$) and writes observations, rewards, etc... A policy is also an agent reading observations, and writing actions . In the supervised learning setting, a dataloader is an \texttt{Agent} that writes data and labels into the \texttt{Workspace} and a predictive model is reading a datapoint and writing a prediction, etc.  Considering that everything is an Agent has two advantages: i) the programmer only has to know what an agent is to implement complex models or understand existing codes. For instance, it avoids having one specific API for environments and one other for policies. ii) It allows the use of parametric agents everywhere. For instance, a \texttt{GymAgent} wrapping an openAI Gym environment can be replaced at any time by a parametric agent (e.g a world model) without changing the rest of the code, leading to the easy development of model-based RL methods. 

\textbf{Concretely,} it means that understanding how \SALINA{} works just necessitates knowing how to manipulate a workspace, and how to define an agent (see next sections). With these two concepts in mind, you are ready to develop any sequential learning model and will be able to execute it on multiple CPUs and GPUs.

\subsection*{Advantages of the Library}

Given these two \SALINA{} principles, let us now present the good properties of such a library:
\begin{itemize}
    \item \textbf{Simplicity: } Understanding the \texttt{Agent} and \texttt{Workspace} API is enough to understand \SALINA{} and to implement complex sequential decision models. There are no hidden mechanisms and the two classes are very simple and familiar to any PyTorch user.
    \item \textbf{Modularity: } \SALINA{} allows one to build complex agent by combining simpler ones by using pre-defined container agents.
    \item \textbf{Flexibility: } \SALINA{} provides additional tools to facilitate the implementation of complex models. \SALINA{} comes with wrappers capturing openAI Gym environments as agents, DataLoader as agents, and Brax environments as agents, allowing one to quickly develop a large variety of models. Moreover, with the ability to replay agents on a workspace, there is no need to have particular replay buffer implementations and it is very easy to do batch RL by reading a workspace directly from a dataset. 
    \item \textbf{Scaling: } \SALINA{} provides a \texttt{NRemoteAgent} wrapper that can execute any agent over multiple processes, speeding-up the computation of any particular agent. Used in addition to the possibility of having agents on CPU or GPU, it make the library able to scale to very large problems, with only few modifications to the code.
\end{itemize}

 As of today, \SALINA{} includes implementations of different algorithms that can be used to understand the principles of the library: Double DQN \cite{vanhasselt2015deep}, DDPG \cite{lillicrap2019continuous}, PPO \cite{schulman2017proximal}, TD3 \cite{fujimoto2018addressing}, A2C \cite{mnih2016asynchronous}, Reinforce and Behavioral Cloning . These algorithms have been designed to be easily modified, are all agnostic to the architecture of the policy (e.g recurrent policies) and make use of both multiple CPUs and GPUs. We also provide an implementation of cascade models in computer vision (using a sequence of spatial transformers) to help you understand how \SALINA{} can handle supervised learning settings. More examples will be added in the next months.

\section{More about Agents}

\subsection*{Controlling Agents: } \SALINA{} allows one to control the execution of agents through the extra arguments provided in the forward method. (see Figure \ref{code:controlagent}). This allows us to control the behavior of an agent at execution time, for instance by specifying if we want to execute policies in stochastic or deterministic mode, the value of $\epsilon$ for $\epsilon$-greedy policies, etc.. It can be also used to switch between step-by-step computation or global computation when using recurrent neural networks, etc.
As a useful abstraction, the \texttt{TAgent} class forces the use of a $t$ arguments when defining an agent, $t$ representing a particular timestep.  It is used as a base class when defining agents that operate on a particular timestep (e.g environment, policy, etc.)

\subsection*{Combining Agents}

Like PyTorch modules, \SALINA{} agents can be combined to build more complicated agents. We provide two container agents:
\begin{itemize}
    \item \texttt{Agents(agent1,agent2,agent3,...)} executes the different agents the one after the other (see Figure \ref{code:tagents}).
    \item \texttt{TemporalAgent(agent)} executes an agent (e.g a \texttt{TAgent}) over multiple timesteps in the workspace, or until a given condition is reached (see Figure \ref{code:tempagent}).
\end{itemize}

Examples of use are given in Figure \ref{code:agents}.

\subsection*{Useful provided Agents}

We provide a (growing) list of implemented agents to simplify the implementation of algorithms: 
\begin{itemize}
\item  \texttt{GymAgent} (resp. \texttt{AutoResetGymAgent}) are agents able to execute a batch of gym environments  \cite{gym} without (resp. with) auto-resetting. These agents produce multiple variables in the workspace: 'env/env\_obs', 'env/reward', 'env/timestep', 'env/done', 'env/initial\_state', 'env/cumulated\_reward', ... When called at timestep \texttt{t=0}, then the environments are automatically reset. At timestep \texttt{t>0}, these agents will read the 'action' variable in the workspace at time $t-1$
\item \texttt{BraxAgent} is quite similar to gym agents, but using the brax environment \cite{brax}. 
\item \texttt{DataLoaderAgent} is an agent based on a torch \texttt{Dataset} used to load a data into a workspace (e.g for supervised learning)
\end{itemize}

\subsection*{Replaying Agents}

One interesting property of \SALINA{} is that it naturally allows one to replay agents over a workspace. For instance, let us consider that we are doing off-policy RL, where an agent $\pi_1 $ is used to sample trajectories, and another agent $\pi_2$ is used to compute some statistics or loss function on collected trajectories. Such a setting can be easily implemented by first executing the environment agent and the agent $\pi_1$ together over a workspace (to generate a trajectory). In a second time, it is possible to execute the second agent $\pi_2$ over the same workspace: the second agent has access to all the variables previously written by the environment agent (e.g observation , reward, etc.) and will overwrite the variables written by the first policy agent, or eventually add new variables (see Figure \ref{code:replay}). In addition of code clarity improvement, not replaying the environment agent also saves computation time. 

\subsection{Discussion}

The objective of the \SALINA{} API is to remain as simple as possible, with as few abstractions as possible. As in PyTorch, we will incrementally grow the list of agents that we provide to facilitate the implementation of algorithms, depending on the demand from the users.

\section{Scaling with SaLinA}

\SALINA{} provide two ways to scale any algorithm: i) use of multiple CPUs and ii) use of multiple GPUs. Using multiple CPUs and/or GPUs is achieved with minimal modifications to your code.

\subsection{Using GPUs}

The use of GPUs in \SALINA{} is as simple as the use of GPUs in PyTorch. Each agent can be moved to a particular device using the \texttt{Agent.to(device)} method. The workspace is 'device agnostic' and can store tensors in any/multiple devices. Note that, if needed, a new workspace containing copies of the tensors on a specified device can be built through \texttt{workspace.to(device)}. 

\subsection{Parallel Execution of Agents}

\SALINA{} allows one to execute agents in different processes speeding up their execution. Moreover, it makes it possible to execute agents without blocking the other agents, for instance when performing an asynchronous evaluation of RL policies while training, or when implementing asynchronous algorithms like IMPALA. 

The remote agents in \SALINA{} make use of a \textbf{shared workspace} where tensors are in shared memory allowing fast computation. Note that agents in multiple processes are executed in \texttt{torch.no\_grad} mode. 

The way the paralellization is acheived is as follows:
\begin{itemize}
    \item First, a single agent is defined by the user.
    \item Then this agent is executed over an empty workspace, allowing to capture the size of the tensors that will be written by this agent. All these tensors have a batch dimension of size $B$. 
    \item A new workspace is built in shared memory, where all the tensors have the same dimension, except the batch dimension that will be of size $B \times n$, $n$ being the number of processes that will be created. 
    \item Then, the single agent is copied over $n$ different processes, each process reading and writing only $B$ of the $B\times n$ batch dimension of the workspace's tensors.
\end{itemize}

These dfferent steps are made automatic by using the \texttt{NRemoteAgent.create} function that takes in charge the described process. Note that, once the \texttt{NRemoteAgent} has been created, it can be used exactly like a normal agent. 

This process is illustrated in Figure \ref{code:remote} and in Figure \ref{code:aremote} for non-blocking execution. The RL algorithms in the repository also provide good examples of how multiple processes agents can be defined and used.

\section{Additional Remarks}

\paragraph{Speed: } One usual remark when facing a new library is about the computation speed. \SALINA{} is a full python library that introduces very few overheads and provides a computation speed comparable to existing alternatives.  

\paragraph{From policies to recurrent policies: } The workspace mechanism allows implementing complex policies easily, without changing anything in the rest of the code. As an example, we show in Figure \ref{code:nrec} and Figure \ref{code:rec} how one can moves from a simple policy to a recurrent one in few lines without modifying the learning algorithm, by just changing the agents involved in the computation.

\paragraph{Replay Buffer: } In \SALINA{}, there is no real need to implement a complex replay buffer class since a set of workspaces can naturally be used as a replay buffer through the replay capability of the agents (for instance by using a list of workspaces, each workspace capturing a batch of trajectories). When one wants to perform more complex operations on the replay buffer, we also provide a \texttt{salina.rl.ReplayBuffer} class in \SALINA{} to facilitate high-level operations (e.g splitting trajectories in sliding windows, etc.)

\paragraph{Batch RL:} (Figure \ref{code:batchrl}) A workspace can also be directly build from a dataset (see \texttt{salina\_examples/bc}) of saved trajectories. By using the replay capability of \SALINA{}, it is easy to compute complex losses over fixed trajectories to obtain behavioral cloning for instance. Alternating between acquired trajectories and demonstrations is also very simple.

\paragraph{Model-based RL: } (Figure \ref{code:modelrl}) Since environments in \SALINA{} are agents, it is possible to replace, at any time, any environment agent by an agent that models the world. By replacing an environment with a parametric agent, the gradient will naturally flow over the environment, allowing fast learning. Note that it also allows training the parameters of the world model. 

\paragraph{Multi-agent RL: } (Figure \ref{code:multirl}) Everything is an agent, so implementing multi-agent settings in \SALINA{} is natural by combining multiple agents in a single one.

\paragraph{The \SALINA{} RL benchmark: } In addition to the core library, we also provide a set of implemented and benchmarked algorithms in the reinforcement learning domain. As of today, \SALINA{} includes implementations of Double DQN \cite{vanhasselt2015deep}, DDPG \cite{lillicrap2019continuous}, PPO \cite{schulman2017proximal}, TD3 \cite{fujimoto2018addressing}, A2C \cite{mnih2016asynchronous}, Reinforce and Behavioral Cloning. These algorithms have been designed to be easily modified, and are all agnostic to the architecture of the policy.

\section{Conclusion}

By combining agents, \SALINA{} offers a new way to implement sequential decision-making algorithms. It is a lightweight library (few hundred of lines of codes), that is very flexible and that works at scale. It allows one to imagine new algorithms and to easily test new ideas without sacrificing training and testing speed. On the personal side, we really like this library, and expect that it will allow you to make your own ideas concrete.

Future perspectives include: a) allowing the execution of agents on remote computers b) development of new tools agent to facilitate implementation c) and the building of a zoo of algorithms in different domains.

\section*{Acknowledgements} We would like to acknowledge some of our colleagues that help us to finalize the library , the documentation and the white paper. In random order: Shagun Sodhani, Brian Vaughan, Alessandro Lazaric, Nicolas Usunier, Jonas Gehring, Maxim Grechkin, Donny Greenberg, Denis Yarats, Vikash Kumar, Vincent Moens, Xiaomeng Yang, Jason Gaucci, ... 

\newpage

\bibliography{references}
\bibliographystyle{alpha}

\newpage
\section{Code Examples}
\label{section:examples}

\begin{figure}[h!]
\begin{python}
from salina import Workspace

#Building a workspace that will auto-size to the variables that it will contain
workspace=Workspace()

# Setting a variable 'x' at time t=5 considering a batch size of 4
workspace.set("x",t=5,torch.randn(4,3))

# Getting the value of a variable 'y' at time t=3
workspace.get("y",t=3)

# Setting a variable of shape 4x6 over 12 timesteps
workspace.set_full("loss",torch.randn(12,4,6))

# returns a TxBx... variable (6x4x3 in our example)
workspace["x"]

# Moving the workspace to gpu
workspace=workspace.to("cuda:0")

\end{python}
\caption{Example of function to manipulate a \texttt{salina.Workspace}}
\label{code:workspace}
\end{figure}

\begin{figure}[h!]
\begin{python}
from salina import Agent

class LinearAgent(Agent):
    def __init__(self,n_input,n_output):
        super().__init__()
        self.model=nn.Linear(n_input,n_output)
    
    def forward(self,t,**args):
        x=self.get(("x",t))
        y=self.model(x)
        self.set(("y",t))

\end{python}
\caption{Example of definition of a new agent computing a linear transformation over variable 'x' at time 't', writing variable 'y' at time t}
\label{code:linearagent}
\end{figure}

\begin{figure}[h!]
\begin{python}
from salina import Agent

class CrossEntropyAgent(Agent):
    def __init__(self):
        super().__init__()
    
    def forward(self,**args):
        y=self.get("y")
        predicted_y=self.get("predicted_y")
        loss=F.cross_entropy(predicted_y,y)
        self.set("loss",loss)

\end{python}
\caption{Example of definition of a new agent that computes a cross entropy loss over all timesteps between 'predicted\_y' and 'y'}
\label{code:lossagent}
\end{figure}

\begin{figure}[h!]
\begin{python}

agent=MyAgent(...)
workspace=Workspace()

# Execute the agent over the workspace
agent(workspace)

\end{python}
\caption{Example of use of an agent}
\label{code:executeagent}
\end{figure}

\begin{figure}[h!]
\begin{python}

from salina import Agent

class FillAgent(Agent):
    def __init__(self):
        super().__init__()
    
    def forward(self,var_name,value,n_steps,**args):
        for t in range(n_steps):
            self.set((var_name,t),torch.tensor([value])

my_agent=FillAgent()
workspace=Workspace()
my_agent(workspace,var_name="x",value=1.0,n_steps=100)

\end{python}
\caption{Example of a simple agent that fills a variable with a particular value. Extra arguments for this agent can be used when executing the agent over a workpace, allowing one to control this agent. }
\label{code:controlagent}
\end{figure}

\begin{figure}[h!]
\begin{python}
class Agents(Agent):
    def __init__(self, *agents,name=None):
        super().__init__(name=name)
        for a in agents:
            assert isinstance(a, Agent)
        self.agents = nn.ModuleList(agents)

    def __call__(self, workspace, **args):
        for a in self.agents:
            a(workspace, **args)
\end{python}
\caption{The definition of the \texttt{Agents} class in \SALINA{}. An \texttt{Agents} just execute a list of agents the one after the other. }
\label{code:tagents}
\end{figure}

\begin{figure}[h!]
\begin{python}
class TemporalAgent(Agent):
    def __init__(self, agent,name=None):
        super().__init__(name=name)
        self.agent = agent

    def __call__(self, workspace, t=0, n_steps=None, stop_variable=None,**args):
        _t=t
        while True:
            #Execute the agent at time t
            self.agent(workspace, t=_t, **args)
            
            # Break is stop condition is reached
            if not stop_variable is None:
                s = workspace.get(stop_variable, _t)
                if s.all():
                    break
            _t+=1
            
            #Break if n_steps is reached
            if not n_steps is None:
                if _t>=t+n_steps:
                    break
\end{python}
\caption{The definition of the \texttt{TemporalAgent} class in \SALINA{}. A \texttt{TemporalAgent} executes an agent from time $t$ to time $t+n\_steps$. It may stop earler if a particular condition is reached}
\label{code:tempagent}
\end{figure}

\begin{figure}[h!]
\begin{python}

from salina import Agent
agent1=MyAgent(...)
agent2=OtherAgent(...)
agent=Agents(agent1,agent2)

# Execute agent1 then agent2
agent(workspace) 

# Extra arguments are transmitted to all the agents
agent(workspace,arg1=...,arg2...)

temporal_agent=TemporalAgent(agent)

#Execute the 'agent' with arguments t=0 to t=49
temporal_agent(workspace,t=0,n_steps=50)

\end{python}
\caption{Combining two agents with \texttt{Agents} and \texttt{TemporalAgent}}
\label{code:agents}
\end{figure}

\begin{figure}[h!]
\begin{python}

env=GymAgent(....,n_envs=16) # batch of 16 environments
acquisition_policy=My_Acquisition_Policy_Agent(...)

policy=PolicyAgent(...)

acquisition_agent=TemporalAgent(Agents(env,acquisition_policy))
workspace=Workspace()

# Acquisition of a batch of trajectories
acquisition_agent(workspace,t=0,n_steps=100) 

# Execution of the policy over the acquired trajectories
policy(workspace,t=0,n_steps=100)

\end{python}
\caption{Replay of agents: The execution of $policy$ over the workspace will potentially overwrite variables written by the acquisition policy, or even create new variables in the workspace}
\label{code:replay}
\end{figure}

\begin{figure}[h!]
\begin{python}

agent=MyAgent(...)
execution_args={....}
# Create 4 copies of the agent in 4 different processes.
# It also returns a remote workspace to use with the remote agent
remote_agent,remote_workspace=NRemoteAgent.create(agent,num_processes=4, **execution_args)

remote_agent(remote_workspace,**execution_args)

#If one wants to convert the remote workspace to a normal workspace
#It is usefull when executing new agents over the workspace (e.g replay)
workspace=Workspace(remote_workspace)

\end{python}
\caption{Creating and using an agent over multiple processes}

\label{code:remote}
\end{figure}

\begin{figure}[h!]
\begin{python}

agent=MyAgent(...)
execution_args={....}
# Create 4 copies of the agent in 4 different processes.
# It also returns a remote workspace to use with the remote agent
remote_agent,remote_workspace=NRemoteAgent(agent,num_processes=4, **execution_args)

remote_agent._asynchronous_call(remote_workspace,**execution_args)

while remote_agent.is_running():
    #Do watever you want

#Here, the remote_workspace is completed and you can use it
workspace=Workspace(remote_workspace)

\end{python}
\caption{Using a remote agent in non blocking model}
\label{code:aremote}
\end{figure}

\begin{figure}[h!]
\begin{python}

class Policy(TAgent):
    def __init__(self):
        self.model=nn.Linear(...,...)
    
    def forward(self,t,**args):
        observation=set.get(("env/env_obs",t))
        scores=self.model(observation)
        probabilities=torch.softmax(scores,dim=-1)
        action=torch.distributions.Categorical(probabilities).sample()
        self.set(("action",t),action)

env_agent=GymAgent(.....)
policy_agent=Policy(...)

learn(env_agent,policy_agent)
\end{python}
\caption{Illustration of the classical RL workflow in \SALINA{}}
\label{code:nrec}
\end{figure}

\begin{figure}[h!]
\begin{python}

class Policy(TAgent):
    def __init__(self):
        self.model=nn.Linear(...,...)
    
    def forward(self):
        observation=set.get(("z",t))
        scores=self.model(observation)
        probabilities=torch.softmax(scores,dim=-1)
        action=torch.distributions.Categorical(probabilities).sample()
        self.set(("action",t),action)

class RecAgent(TAgent):
    def __init__(self):
        self.rec_model=...
    
    def forward(self,t,**args):
        observation=set.get(("z",t))
        batch_size=observation.size()[0]
        if t==0:
            self.set(("z",t),torch.zeros(B,self.N))
        else:
            z=self.get(("z",t-1))
            z=self.rec_model(z,observation)
            self.set(("z",t),z)

env_agent=GymAgent(.....)
policy_agent=Agents(RecAgent(...),Policy(...))

learn(env_agent,policy_agent)
\end{python}
\caption{Switching to a  recurrent agent, without modify the learning algorithm (see Figure \ref{code:nrec} for comparison)}
\label{code:rec}
\end{figure}

\begin{figure}[h!]
\begin{python}

# ==== Classical RL setting
env=GymAgent(...)
policy=PolicyAgent(...)
agent=TemporalAgent(Agents(env,policy))
workspace=Workspace()

while True:
    agent(workspace,t=0,n_steps=100) # The agent does trajectories acqusition
    loss=compute_loss(workspace)
    ....

# === Batch RL setting

policy=PolicyAgent(...)
workspace=Workspace()

while True:
    workspace=dataset.read_workspace() # Trajectories come from a dataset
    policy(workspace,t=0,n_steps=100) # The policy is replayed among trajectories
    loss=compute_loss(workspace)
    ....

\end{python}
\caption{Batch RL principles in \SALINA{}. The workspace is read from a dataset. The policy is then replayed on the workspace to compute the loss.}
\label{code:batchrl}
\end{figure}

\begin{figure}[h!]
\begin{python}

# ==== Classical RL setting
env=GymAgent(...)
policy=PolicyAgent(...)
agent=TemporalAgent(Agents(env,policy))
workspace=Workspace()

while True:
    agent(workspace,t=0,n_steps=100)
    loss=compute_loss(workspace)

# === Model-based setting

env = MyParametricEnv(...)
policy=PolicyAgent(...)
agent=TemporalAgent(Agents(env,policy))
workspace=Workspace()

while True:
    agent(workspace,t=0,n_steps=100)
    loss=compute_loss(workspace)

\end{python}
\caption{Model-based RL principles in \SALINA{}. It can be done by just switching the environment agent by a parameterized agent written in PyTorch. Gradient will naturally flow through the environment.}
\label{code:modelrl}
\end{figure}

\begin{figure}[h!]
\begin{python}

# ==== Multiagent RL setting
env=GymAgent(...)
policy_1=PolicyAgent(...)
policy_2=PolicyAgent(...)
agent=TemporalAgent(Agents(env,policy_1,policy_2))
workspace=Workspace()

while True:
    agent(workspace,t=0,n_steps=100)
    loss=compute_loss(workspace)
    
    #The gradient of the loss is back-propagated over agent 1 and agent 2

\end{python}
\caption{Multi-agent RL principles in \SALINA{}. }
\label{code:multirl}
\end{figure}

\end{document}